\documentclass[11pt]{article}

\usepackage[]{EMNLP2023}

\usepackage{times}
\usepackage{latexsym}

\usepackage[T1]{fontenc}
\usepackage[utf8]{inputenc}

\usepackage{microtype}

\usepackage{inconsolata}
\usepackage{CJKutf8}
\usepackage{graphicx}
\usepackage{amssymb}
\usepackage{multirow}
\usepackage{pifont}
\usepackage{tabu}
\usepackage{booktabs}
\usepackage{color}
\usepackage{hyperref}

\newcommand{\cmark}{\ding{51}}%
\newcommand{\xmark}{\ding{55}}%
\title{Learning to Compose Representations of Different Encoder Layers towards Improving Compositional Generalization}

\author{Lei Lin\textsuperscript{1,2\footnotemark[1]}~~~
Shuangtao Li\textsuperscript{1,2\footnotemark[1]}~~~
Yafang Zheng\textsuperscript{1,2\thanks{~~Equal Contribution.}}~~~
Biao Fu\textsuperscript{1,2}~~~
Shan Liu\textsuperscript{1,2}~~~\\
\textbf{
Yidong Chen\textsuperscript{1,2}~~~
Xiaodong Shi\textsuperscript{1,2\thanks{~~Corresponding Author.}}~~~}\\
\textsuperscript{1}Department of Artificial Intelligence, School of Informatics, Xiamen University\\
\textsuperscript{2}Key Laboratory of Digital Protection and Intelligent Processing of Intangible Cultural Heritage\\ of Fujian and Taiwan (Xiamen University), Ministry of Culture and Tourism, China\\
\texttt{\{linlei\}@stu.xmu.edu.cn,\{ydchen,mandel\}@xmu.edu.cn} \\}

\begin{document}
\maketitle
\begin{abstract}
Recent studies have shown that sequence-to-sequence (seq2seq) models struggle with compositional generalization (CG),
% are limited in solving the compositional generalization (CG) tasks, 
i.e., the ability to systematically generalize to unseen compositions of seen components. There is mounting evidence that one of the reasons hindering CG is the representation of the encoder uppermost layer is entangled, i.e., the syntactic and semantic representations of sequences are entangled. 
% In other words, the syntactic and semantic representations of sequences are twisted inappropriately. 
% However, most previous studies mainly concentrate on enhancing semantic information at token-level, rather than composing the syntactic and semantic representations of sequences appropriately as humans do. 
% However, we consider the representation entanglement problem they found is not comprehensive, and further hypothesize that source keys and values representations passing into different decoder layers are also entangled. 
% However, we consider that the previously identified representation entanglement problem is not comprehensive enough. In other words, we hypothesize that the representations of source keys and values passing into different decoder layers are also entangled.
However, we consider that the previously identified representation entanglement problem is not comprehensive enough. Additionally, we hypothesize that the source keys and values representations passing into different decoder layers are also entangled.
Starting from this intuition, we propose \textsc{CompoSition} (\textbf{Compo}se \textbf{S}yntactic and Semant\textbf{i}c Representa\textbf{tion}s), an extension to seq2seq models which learns to compose representations of different encoder layers dynamically for different tasks, since recent studies reveal that the bottom layers of the Transformer encoder contain more syntactic information and the top ones contain more semantic information. Specifically, we introduce a \textit{composed layer} between the encoder and decoder to compose different encoder layers' representations to generate specific keys and values passing into different decoder layers.
\textsc{CompoSition} achieves competitive results on two comprehensive and realistic benchmarks, which empirically demonstrates the effectiveness of our proposal. Codes are available at~\url{https://github.com/thinkaboutzero/COMPOSITION}.
% Through comprehensive evaluations on semantic parsing and machine translation benchmarks, \textsc{CompoSition} achieves competitive and even \textbf{sate-of-the-art} results, demonstrating the empirical effectiveness of our proposal.
\end{abstract}

\section{Introduction}
\label{sec:intro}

% The principle of compositionality is a crucial property of language: the meaning of a complex expression is fully determined by its syntactic structure (syntactic information) and the meanings of its constituents (semantic information)~\cite{pelletier2003context-tpoc,szabo2004compositionality-tpoc}.  Based on this principle, humans exhibit \textit{compositional generalization} (CG) when they use language, i.e., 
A crucial property of human language learning is its \textit{compositional generalization} (CG) --- the algebraic ability to understand and produce a potentially infinite number of novel combinations from known components~\cite{fodor1988connectionism,lake2017building}. For example, if a person knows ``the doctor watches a movie'' [\begin{CJK*}{UTF8}{gbsn}医生看电影\end{CJK*}]\footnote{The sentence in ``[]'' denotes the Chinese translation.} and ``the lawyer'' [\begin{CJK*}{UTF8}{gbsn}律师\end{CJK*}], then it is natural for the person to know the translation of ``the lawyer watches a movie'' is [\begin{CJK*}{UTF8}{gbsn}律师看电影\end{CJK*}] even though they have never seen it before. Such nature is beneficial for generalizing to new compositions of previously observed elements, which is often required in real-world scenarios.

\begin{figure}[!t]
    \centering
    \includegraphics[width=0.95\linewidth]{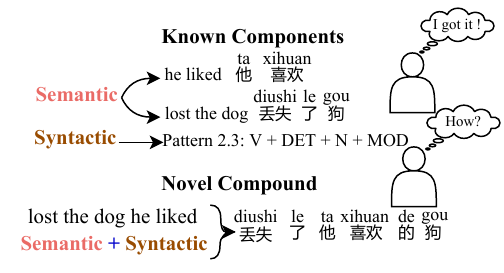}
    \caption{Examples from CoGnition~\cite{li2021compositional} show the workflow of how humans exhibit CG. Suppose interpreters know the translation: [\begin{CJK*}{UTF8}{gbsn}丢失了狗\end{CJK*}] for ``lost the dog'' and [\begin{CJK*}{UTF8}{gbsn}他喜欢\end{CJK*}] for ``he liked'' (semantic information). When they first encounter ``lost the dog he liked'', they can correctly translate [\begin{CJK*}{UTF8}{gbsn}丢失了他喜欢的狗\end{CJK*}] instead of [\begin{CJK*}{UTF8}{gbsn}丢失了狗他喜欢\end{CJK*}] depending on Pattern 2.3 (syntactic information).}
    \label{fig:fig1}
\end{figure}
Despite astonishing successes across a broad range of natural language understanding and generation tasks~\cite{sutskever2014sequence,dong2016language,vaswani2017attention}, neural network models, in particular the very popular sequence-to-sequence (seq2seq) architecture, are argued difficult to capture the compositional structure of human language~\cite{lake2018generalization,keysers2019measuring,li2021compositional}. A key reason for failure on CG is different semantic factors (e.g., lexical meaning and syntactic patterns) required by CG are entangled, which was proved explicitly or implicitly to exist in the representation of the encoder uppermost layer (encoder entanglement problem) by previous studies~\cite{li2019compositional,raunak2019compositionality,russin2019compositional,liuqian2020compositional,DBLP:conf/acl/LiuALLCLWZZ21,JiangB21indu,zheng2021disentangled,yin2022categorizing,corrabs220210745,corrabs221001603,eaclCazzaroLQC23}. In other words, the syntactic and semantic representations of sequences are entangled.

% Therefore, one natural question can be raised: can we train the model to learn to compose the syntactic and semantic representations of sequences appropriately. We get insights from the perspective of how humans exhibit CG. As shown in Figure~\ref{fig:fig1}, suppose interpreters know the translation: [\begin{CJK*}{UTF8}{gbsn}丢失了狗\end{CJK*}] for ``lost the dog'' and [\begin{CJK*}{UTF8}{gbsn}他喜欢\end{CJK*}] for ``he liked'' (semantic information). When they first encounter ``lost the dog he liked'', they can correctly translate [\begin{CJK*}{UTF8}{gbsn}丢失了他喜欢的狗\end{CJK*}] instead of [\begin{CJK*}{UTF8}{gbsn}丢失了狗他喜欢\end{CJK*}] depending on Pattern 2.3 (syntactic information) of Figure~\ref{fig:fig1} (see Figure~\ref{fig:fig1}).\footnote{The postpositive modifier translation from English to Chinese requires word reordering.} It is clear that the key to CG ability of humans is an appropriate composition of semantic and syntactic information.
% Thus, one intuitive solution is to first disentangle the entangled representation into two separate syntactic and semantic representations (\textbf{disentanglement}), and then compose them appropriately (\textbf{composition}).
% Inspired by humans受到人类...启发，一条研究路线在于如何利用句法和语义信息来缓解编码器表示耦合问题，从而提升模型的组合泛化性能。然而，我们认为...
In order to alleviate the encoder entanglement problem, one line of research on CG mainly concentrate on improving the encoder representation or separating the learning of syntax and semantics which adopt similar approaches to humans’ strategies for CG (see Figure~\ref{fig:fig1}).
% , although their starting points are not derived from it. 
% As shown in Fig.~\ref{fig:fig1}, suppose interpreters know the translation: [\begin{CJK*}{UTF8}{gbsn}丢失了狗\end{CJK*}] for ``lost the dog'' and [\begin{CJK*}{UTF8}{gbsn}他喜欢\end{CJK*}] for ``he liked'' (semantic information). When they first encounter ``lost the dog he liked'', they can correctly translate [\begin{CJK*}{UTF8}{gbsn}丢失了他喜欢的狗\end{CJK*}] instead of [\begin{CJK*}{UTF8}{gbsn}丢失了狗他喜欢\end{CJK*}] depending on Pattern 2.3 (syntactic information) of Fig.~\ref{fig:fig1}.\footnote{The postpositive modifier translation from English to Chinese requires word reordering.} 
% It is clear that the key to CG ability of humans is \textbf{an appropriate composition of semantic and syntactic information}.
Specifically, several works either produce two separate syntactic and semantic representations, and then compose them~\cite{li2019compositional,russin2019compositional,JiangB21indu} or design external modules, and then employ a multi-stage generation process~\cite{liuqian2020compositional,DBLP:conf/acl/LiuALLCLWZZ21,corrabs220210745,corrabs221001603,eaclCazzaroLQC23}.
% to achieve the same goal where one maps the input token into the output and the other aligns input to output tokens. 
Moreover, some studies explore bag-of-words pre-training~\cite{raunak2019compositionality}, newly decoded target context~\cite{zheng2021disentangled,zheng2022real} or prototypes of token representations over the training set~\cite{yin2022categorizing} to improve the encoder representation. 
% However, we further hypothesize 
Furthermore, we hypothesize that the source keys and values representations passing into different decoder layers are also entangled (keys, values entanglement problem), not just the representation of the encoder uppermost layer. We will further illustrate it in Section~\ref{sec:kvep}.

Therefore, one natural question can be raised: how to alleviate keys, values entanglement problem.
% Instead of separating the learning of syntax and semantics like previous work, we examine CG from a new perspective to solve it, i.e., utilizing different encoder layers' information, since we conduct preliminary analysis provided in Appendix A, and concluded that the bottom layers of the Transformer encoder contain more syntactic information and the top ones contain more semantic information.
As a remedy, we examine CG from a new perspective to solve it, i.e., utilizing different encoder layers' information. We conduct preliminary analysis provided in Appendix~\ref{sec:apa}, and conclude that the bottom layers of the Transformer encoder contain more syntactic information and the top ones contain more semantic information. Inspired by this, we collect representations outputted by each encoder layer instead of separating the learning of syntax and semantics. So one intuitive solution to solve keys, values entanglement problem is to learn different and specific combinations of syntactic and semantic information (i.e., representations outputted by each encoder layer) for keys and values of different decoder layers. We argue that \textit{an effective composition is to provide different combinations for different tasks and a specific combination for a particular task}. For example, the model can learn preference of layers in different levels of the encoder for different tasks (i.e., For A task, the information at encoder layer 0 may be more important, however, for B task, the information at encoder layer 5 may be more important). Additionally, the model can select which encoder layer of information is most suitable for itself (that is, which encoder layer of information is the most important) for a particular task. Inspired by that, we propose \textit{the composed layer} (learnable scalars or vectors) to generate different specific source keys and values passing into different decoder layers for different particular tasks, since we argue that the learned scalars or vectors (i.e., different dynamic composition modes) by the model itself during training process can be dynamically adjusted for different particular tasks, and provide a way to learn preference of layers in different levels of the encoder for a particular task. Putting everything together, we propose \textsc{CompoSition} (\textbf{Compo}se \textbf{S}yntactic and Semant\textbf{i}c Representa\textbf{tion}s), an extension to seq2seq models that learns to compose the syntactic and semantic representations of sequences dynamically for different tasks. \textsc{CompoSition} is simple yet effective, and mostly applicable to any seq2seq models without any dataset or task-specific modification.
Experimental results on CFQ~\cite{keysers2019measuring} (semantic parsing) and CoGnition~\cite{li2021compositional} (machine translation, MT) empirically show that our method can improve generalization performance, outperforming competitive baselines and other techniques. Notably, \textsc{CompoSition} achieves \textbf{19.2\%} and \textbf{50.2\%} (about \textbf{32\%}, \textbf{20\%} relative improvements) for instance-level and aggregate-level error reduction rates on CoGnition. Extensive analyses demonstrate that composing the syntactic and semantic representations of sequences dynamically for different tasks leads to better generalization results.
% , outperforming competitive baselines and more specialized techniques.

\section{Related Work}
\label{sec:rw}

\noindent\textbf{Compositional Generalization.} 
% CG has long played a prominent role in language understanding, explaining why we understand novel compositions of previously observed elements. Therefore, 
After realizing existing neural models still struggle in  scenarios requiring CG~\cite{lake2018generalization,keysers2019measuring,li2021compositional}, there have been various studies attempt to improve the model’s ability of CG, including data augmentation~\cite{andreas2020good,DBLP:conf/iclr/AkyurekAA21,DBLP:conf/naacl/YangZY22,li2023learning}, modifications on model architecture~\cite{li2019compositional,russin2019compositional,NyeS0L20,liu2020lane,DBLP:conf/acl/LiuALLCLWZZ21,ZhengL21,HerzigB20,chaabouni2021can,DBLP:conf/iclr/MittalRRBL22,zheng2023layer}, intermediate representations~\cite{furrer2020compositional,herizgunlock21}, meta-learning~\cite{Lake19meta,ConklinWST20}, explorations on pre-trained language models~\cite{furrer2020compositional,zhou2023leasttomost}, auxiliary objectives~\cite{JiangB21indu,yin2023consistency}, two representations~\cite{li2019compositional,russin2019compositional,JiangB21indu} and enriching the encoder representation~\cite{raunak2019compositionality,zheng2021disentangled,zheng2022real,yin2022categorizing,DBLP:conf/emnlp/YaoK22}. One line of research exploring how to alleviate the encoder entanglement problem has attracted much attention. Our work is in line with it, but we examine CG from a new perspective, i.e., utilizing different encoder layers' information. 
% and propose \textsc{CompoSition} to learn to compose the syntactic and semantic information of sequences in different encoder sub-layers appropriately for CG.
% , which is different from previous studies mainly focusing on enhancing semantic information at token-level.

\noindent\textbf{Neural Machine Translation.} Recently, CG and robustness of Neural Machine Translation (NMT) have gained much attention from the research community~\cite{cheng2020advaug,xu2021addressing,lake2018generalization,li2021compositional}, including pre-training~\cite{raunak2019compositionality}, data augmentation~\cite{guo2020sequence}, datasets~\cite{li2021compositional}, and enriching semantic information at token-level~\cite{thrush2020compositional,akyurek2021lexicon,zheng2021disentangled,zheng2022real,yin2022categorizing}. Noteworthily,~\citet{DBLP:conf/acl/DankersBH22} argue that MT is a suitable and relevant testing ground to test CG in
natural language.
% .~\citet{raunak2019compositionality} propose bag-of-words pre-training for the encoder.~\citet{guo2020sequence} propose sequence-level mixup to create synthetic samples. Recently,~\citet{li2021compositional} introduces a practical benchmark dataset called CoGnition which provides qualitative analysis in the forms of fine-grained and systematic composition.
% and observes significant compositional generalization issues. 
% Based on this,~\citet{yin2022categorizing} proposes Proto-Transformer to enhance token-level representations and~\citet{zheng2021disentangled} proposes to adaptively re-encode (at each time step) the source input.  
Different from them, we introduce a composed layer to compose different encoder layers' information dynamically, which is inspired by previous studies about analyzing Transformer~\cite{raganato2018analysis,voita2019bottom}.
% , where the bottom layers of the Transformer contain more syntactic information and the top ones contain more semantic information.

\noindent\textbf{Encoder Layer Fusion.} Encoder layer fusion (EncoderFusion) is a technique to fuse all the encoder layers (instead of the uppermost layer) for seq2seq models, which has been proven beneficial, such as layer attention~\cite{bapna2018training,shen2018dense,wang2019learning}, layer aggregation~\cite{dou2018exploiting,wang2020multi,dou2019dynamic}, and layer-wise coordination~\cite{he2018layer,liu2020layer}. However, other studies show that exploiting low-layer encoder representations fails to improve model performance~\cite{domhan2018much}. 
% \citet{liu2020understanding} consolidates the conflicting conclusions, and proposes SurfaceFusion to only connect the encoder embedding layer to softmax layer. 
% Our method has a similar way to aggregate source information of different encoder layers with previous studies on EncoderFusion. 
The essence of different EncoderFusion works is to explore different ways to combine information from different encoder layers.
Our approach is essentially the same as EncoderFusion work, which explores different ways to combine information from different encoder layers, however, \textit{we propose a new way to combine them}.
Meanwhile, we consider that there are also three distinct differences. \textbf{Firstly}, our method exploits information from all encoder sub-layers and generates specific keys, values passing into different decoder layers while they do not. \textbf{Secondly}, our method shows the effectiveness of utilizing low-layer encoder representations while they have the opposite view (see Appendix~\ref{sec:eler}). \textbf{Thirdly}, we do not share the same motivation or task. Their work focuses on how to transform information across layers in deep neural network scenarios for seq2seq tasks. Our motivation is to compose the syntactic and semantic representations of sequences dynamically for CG.

\section{Methodology}
\label{sec:met}

We adopt the Transformer architecture~\cite{vaswani2017attention} to clarify our method,
% , since it achieves the tremendous successes at various natural language generation and understanding tasks. 
however, \textbf{our proposed method is mostly applicable to any seq2seq models}. In the following, we first introduce the Transformer baseline (Section~\ref{sec:tf}), and then our proposed \textsc{CompoSition} (Section~\ref{sec:cmp}).

\subsection{Transformer}
\label{sec:tf}

The Transformer~\cite{vaswani2017attention} is designed for sequence to sequence tasks which adopts an encoder-decoder architecture. The multi-layer encoder summarizes a source sequence into a contextualized representation and another multi-layer decoder produces the target sequence conditioned on the encoded representation. 

Formally, given a sequence of source sentence $X = \{x_{1}, \ldots ,x_{S} \}$ and a sequence of target sentence $Y = \{y_{1}, \ldots ,y_{T} \}$, where $S, T$ denote the number of source and target tokens, respectively. $\mathcal{D} = \{(X, Y), \ldots \}$ denotes a training corpus, $\mathcal{V}$ denotes the vocabulary of $\mathcal{D}$, and $\theta$ denotes parameters of the Transformer model. The model aims to estimate the conditional probability $p(y_{1}, \ldots ,y_{T} | x_{1}, \ldots ,x_{S})$:
\begin{equation}
    p(Y | X; \theta) = \prod_{t=1}^{T+1} p(y_{t} | y_{< t}, X; \theta),
\end{equation}
where $t$ is the index of each time step, $y_{< t}$ denotes a prefix of $Y$ and each factor $p(y_{t} | X, y_{1}, \ldots ,y_{t-1};$ $\theta)$ is defined as a $softmax$ distribution of $\mathcal{V}$.

\begin{figure}[!t]
    \centering
    \includegraphics[width=0.9\linewidth]{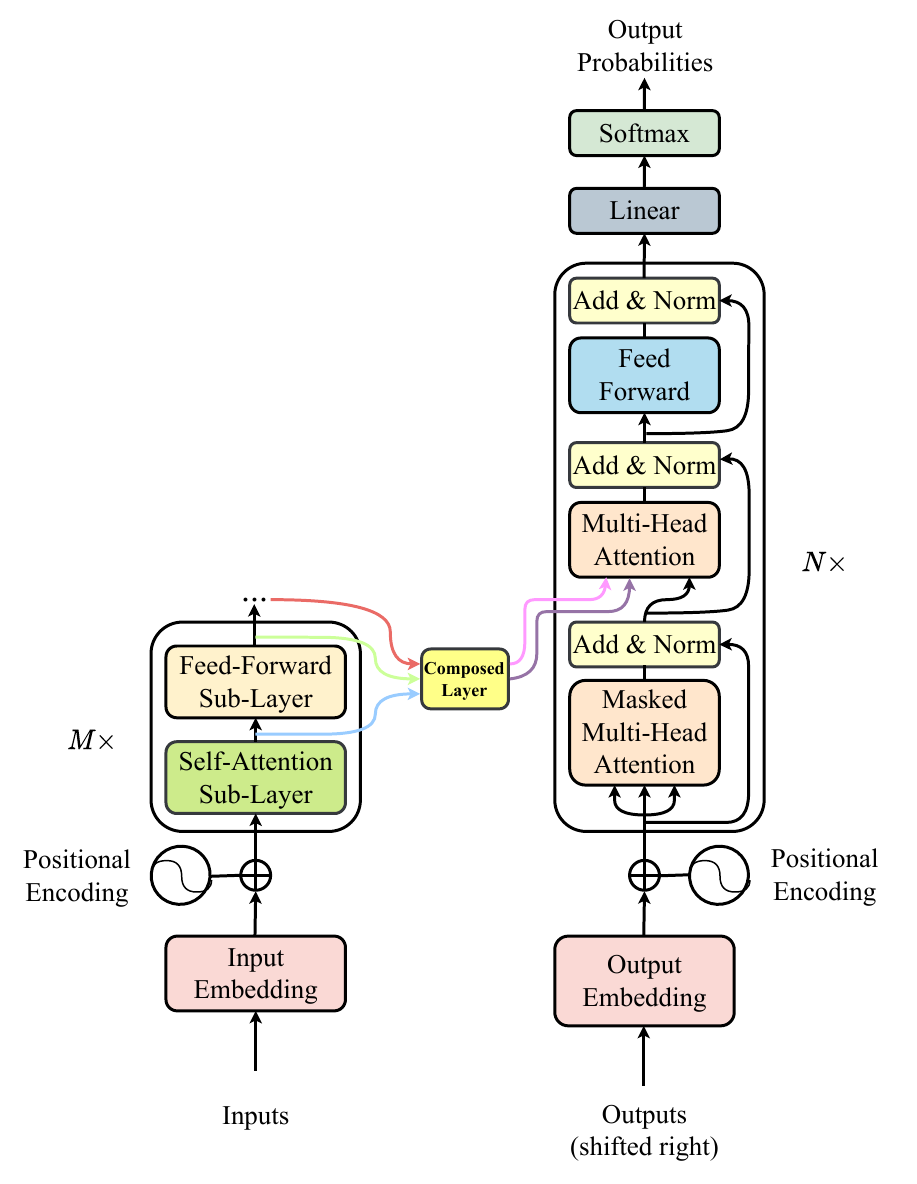}
    \caption{Architecture of \textsc{CompoSition} based on the Transformer. \colorbox{yellow}{The bright yellow block} in the middle denotes the composed layer introduced in Section~\ref{sec:cmp}. \colorbox{red}{The red line} denotes that we collect representations of the same positions for the rest encoder layers.}
    \label{fig:fig2}
\end{figure}
During training, the model are generally optimized with the cross-entropy (CE) loss, which is calculated as follows:
\begin{equation}
    L_{CE}(\theta) = -\sum_{t=1}^{T+1}\log p(y_{t} | y_{< t}, X; \theta).
\end{equation}

During inference, the model predicts the probabilities of target tokens in an auto-regressive mode and generates target sentences using a heuristic search algorithm, such as beam search~\cite{Freitag_2017}.

\subsection{\textsc{CompoSition}}
\label{sec:cmp}

Our proposed \textsc{CompoSition} extends the Transformer by introducing a composed layer between the encoder and decoder.
% , which composes representations outputted by every encoder layer dynamically to generate specific source keys and values passing into different decoder layers for solving keys, values entanglement problem. 
% \textbf{Our method is simple and applicable to a wide range of more realistic tasks} due to little increase in computational and memory cost compared to the Transformer baseline (see Section~\ref{sec:cmc}), which is in contrast to~\cite{raunak2019compositionality,zheng2021disentangled,yin2022categorizing}. 
Figure~\ref{fig:fig2} shows the overall architecture of our approach.

\subsubsection{Composed Layer}
\label{sec:cl}

% To reach the goal of \textbf{disentanglement}, we collect representations outputted by each encoder layer instead of disentangling the representation of the encoder uppermost layer, since they contain less syntactic information. 
% To achieve the goal of \textbf{composition}, 
% We propose a composed layer to compose collected source representations appropriately. Specifically, 
The composed layer is a list consisting of $2N$ learnable vectors due to $2N$ source keys, values passing into $N$ decoder layers, where each vector involves $2M$ learnable scalars or vectors. $M, N$ denote the number of encoder and decoder layers respectively.

\subsubsection{Dynamic Combination}
\label{sec:plc}

Here, we describe how to use the composed layer to compose collected representations dynamically for generating specific keys and values representations passing into different decoder layers. Let $f_{Self-Attention}$ and $f_{Feed-Forward}$ denote a Transformer self-attention sub-layer and feed-forward sub-layer respectively. The embedding layer of the Transformer encoder first maps $X$ to embeddings $H_{0}$, and then $H_{0}$ are fed into a Transformer self-attention sub-layer and feed-forward sub-layer to generate $H_{1}^{SA} \in \mathbb{R}^{d \times S}, H_{1}^{FF} \in \mathbb{R}^{d \times S}$ respectively, where $d$ denotes the hidden size. Next, each subsequent encoder layer takes the previous layer's output as input. The overall process is as follows:
\begin{equation}
    H_{1}^{SA} = f_{Self-Attention}(H_{0}),
\end{equation}
\begin{equation}
    H_{1}^{FF} = f_{Feed-Forward}(H_{1}^{SA}),
\end{equation}
\begin{equation}
    \label{eq5}
    H_{i}^{SA} = f_{Self-Attention}(H_{i-1}^{FF}),
\end{equation}
\begin{equation}
    \label{eq6}
    H_{i}^{FF} = f_{Feed-Forward}(H_{i-1}^{SA}),
\end{equation}
where $2 \leq i \leq M$ denote $i$-th encoder layer. Therefore, we can collect representations outputted by each encoder sub-layer $H_{collect} = \{H_{1}^{SA}, H_{1}^{FF}, \ldots ,H_{M}^{SA}, H_{M}^{FF}\}$. The keys and values of multi-head attention module of decoder layer $l$ are defined to be:
\begin{equation}
    \label{eq7}
    H_{key}^{l} = \sum_{i=1}^{2M}w_{k}^{i}H_{collect}^{i},
\end{equation}
\begin{equation}
    \label{eq8}
    H_{value}^{l} = \sum_{i=1}^{2M}w_{v}^{i}H_{collect}^{i},
\end{equation}
where $w_k^{i} \in \mathbb{R}, w_v^{i} \in \mathbb{R}$ are learnable scalars or vectors and mutually different (e.g. $w_k^{i} \ne w_v^{i}$, $w_k^{i} \ne w_k^{j}$ and $w_v^{i} \ne w_v^{j}$), which weight each collected source representation in a dynamic linear manner.  Eq.~\ref{eq7} and~\ref{eq8} provide a way to learn preference of sub-layers in different levels of the encoder.

\begin{table*}[t]
\centering
\small
\resizebox{1.0\linewidth}{!}{
\begin{tabular}{cccccccc}
\toprule
\multirow{2}{*}{\bf Model} & \multirow{2}{*}{\bf \#Params} & \multicolumn{5}{c}{\bf Compound Translation Error Rate (CTER) $\downarrow$} & \multirow{2}{*}{\bf BLEU $\uparrow$}\\
\cmidrule{3-7}
& & NP & VP & PP & Total & $\Delta$ \\
\midrule
Transformer & 35M & 24.7\%/55.2\% & 24.8\%/59.5\% & 35.7\%/73.9\% & 28.4\%/62.9\% & -/- & 59.5 \\
% \hline
Transformer-Rela & 35M & 30.1\%/58.1\% & 27.6\%/61.2\% & 38.5\%/74.1\% & 32.1\%/64.5\% & +3.7\%/+1.6\% & 59.1 \\
% \hline
Transformer-Small & 25M & 25.1\%/56.9\%  & 25.6\%/60.3\% & 39.1\%/75.0\% & 29.9\%/64.5\% & +1.5\%/+1.6\% & 59.0 \\
% \hline
Transformer-Deep & 40M & 23.3\%/51.6\%  & 24.1\%/58.0\% & 33.8\%/72.6\% & 27.0\%/60.7\% & -1.4\%/-2.0\% & 60.1 \\
% \hline
% Bloom & 7B & 22.8\%/44.0\%  & 39.3\%/68.1\% & 44.7\%/79.6\% & 35.6\%/63.9\% & +7.2\%/+1.0\% & 48.4 \\
\midrule
Bow & 35M & 22.2\%47.9\% & 24.8\%/55.6\% & 	35.0\%/73.2\% & 27.3\%/58.9\% & -1.1\%/-3.0\% & - \\
% \hline
SeqMix & 35M & 24.5\%/49.7\% & 26.9\%/58.9\% & 34.4\%/73.1\% & 28.6\%/60.6\% & +0.2\%/-2.3\% & - \\
% \hline
Dangle & 35M & -/-  & -/- & -/- & 24.4\%/55.5\% & -5.0\%/-7.4\% & 59.7 \\
% \hline
Proto-Transformer & ~42M & 14.1\%/36.5\%  & 22.1\%/50.9\% & 28.9\%/68.2\% & 21.7\%/51.8\% & -6.7\%/-11.1\% & 60.1 \\
Transformer+CReg & 25M & -/- & -/- & -/- & 20.2\%/48.3\% & -8.2\%/-14.6\% & 61.3 \\
$~$R-Dangle$_{\mathrm{sep}}$ & 70M & -/- & -/- & -/- & \bf 16.0\%/42.1\% & \bf -12.4\%/-20.8\% & \bf 63.4 \\
DLCL & 35M & -/- & -/- & -/- & 28.4\%/67.9\% & +0.0\%/+5.0\% & 59.2 \\
\midrule
\textsc{CompoSition} & 35M & \bf 10.0\%/32.6\%  & 22.1\%/54.8\% & 29.2\%/68.5\% & 20.4\%/52.0\% & -8.0\%/-10.9\% & 61.5 \\
% \hline
\textsc{CompoSition}-Rela & 35M & 15.5\%/39.2\%  & 22.4\%/54.0\% & 29.1\%/67.3\% & 22.3\%/53.5\% & -6.1\%/-9.4\% & 61.6 \\
% \hline
\textsc{CompoSition}-Small & 25M & 14.3\%/40.3\%  & 24.4\%/58.1\% & 34.5\%/73.4\% & 24.4\%/57.3\% & -4.0\%/-5.6\% & 60.1 \\
% \hline
\textsc{CompoSition}-Deep & 40M & 11.4\%/34.7\%  & \bf 19.5\%/50.4\% & \bf 26.7\%/65.6\% & 19.2\%/50.2\% & -9.2\%/-12.7\% & 62.0 \\
\bottomrule
\end{tabular}
}
\caption{
CTERs (\%) on CoGnition. We report instance-level and aggregate-level CTERs in the CG-test set, separated by ``/''. In addition, we also report the commonly used metric BLEU score in MT tasks. ``-'' denotes that the results are not provided in the original paper. Results are averaged over 6 random runs.
}
\label{table:t1}
\end{table*}
\section{Experiments}
\label{sec:ex}

We mainly evaluate \textsc{CompoSition} on two comprehensive and realistic benchmarks for measuring CG, including \textit{CFQ}~\cite{keysers2019measuring} and \textit{CoGnition}~\cite{li2021compositional}.
% including semantic parsing (\textit{CFQ}~\cite{keysers2019measuring}) and machine translation (\textit{CoGnition}~\cite{li2021compositional}).

\subsection{Experimental Settings}
\label{sec:est}
\begin{figure}[!t]
    \centering
    \includegraphics[width=1.0\linewidth]{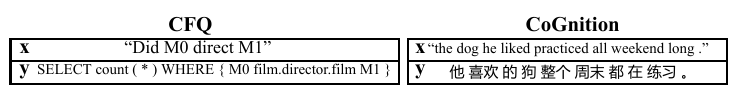}
    \caption{Examples of CFQ and CoGnition.}
    \label{fig:fig3}
\end{figure}

\noindent\textbf{Datasets.} \textit{CoGnition} is a recently released realistic English $\rightarrow$ Chinese (En$\rightarrow$Zh) translation dataset, which is used to systematically evaluate CG in MT scenarios. It consists of a training set of 196,246 sentence pairs, a validation set and a test set of 10,000 samples. In particular, it also has a dedicated synthetic test set (i.e., CG-test set) consisting of 10,800 sentences containing novel compounds, so that the ratio of compounds that are correctly translated can be computed to evaluate the model’s ability of CG directly. 
% The CG-test set consists of 2,160 novel compounds, with up to 5 atoms and 7 words, where novel compounds contains noun phrases (NP), verb phrases (VP) and positional phrases (PP) three most frequent types. Each novel compound is combined with 5 different contexts. 
\textit{CFQ} 
% is a large-scale and realistic dataset consisting of natural language questions paired with its corresponding SPARQL queries against the Freebase knowledge base~\cite{BollackerEPST08}. CFQ 
is automatically generated from a set of rules in a way that precisely tracks which rules (atoms) and rule combinations (compounds) of each example. In this way, we can generate three splits with \emph{maximum compound divergence} (MCD) while guaranteeing a small atom divergence between train and test sets, where large compound divergence denotes the test set involves more examples with unseen syntactic structures. We evaluate our method on all three splits. Each split dataset consists of a training set of 95,743, a validation set and a test set of 11,968 examples.
Figure~\ref{fig:fig3} shows examples of them. 

\noindent\textbf{Data Preprocess.} We follow the same settings of~\citet{li2021compositional} and~\citet{keysers2019measuring} to preprocess CoGnition and CFQ datasets separately. For CoGnition, we use an open-source Chinese tokenizer\footnote{\url{https://github.com/fxsjy/jieba}} to preprocess Chinese and apply Moses tokenizer\footnote{\url{https://github.com/moses-smt/mosesdecoder/blob/master/scripts/tokenizer/tokenizer.perl}} to preprocess English, which is the same in~\citet{lin23leapt} and~\citet{liu2023novel}. We employ byte-pair encoding (BPE)~\cite{sennrichetal2016neural} for Chinese with 3,000 merge operations, generating a vocabulary of 5,500 subwords. We do not apply BPE for English due to the small vocabulary (i.e., 2000). For CFQ, we use the GPT2-BPE tokenizer\footnote{\url{https://github.com/facebookresearch/fairseq/blob/main/examples/roberta/multiprocessing_bpe_encoder.py}} to preprocess source and target English text.

\noindent\textbf{Setup.} For CoGnition and CFQ, we follow the same experimental settings and configurations of~\citet{li2021compositional} and~\citet{zheng2021disentangled} repspectively.
% For CoGnition , we follow the same experimental settings and configurations of~\citet{li2021compositional}. 
% % We build our model on top of Transformer~\cite{vaswani2017attention}. 
% For CFQ, we follow the same experimental settings and configurations of~\citet{zheng2021disentangled}.
% We build our model on top of RoBERTa~\cite{liu2019roberta}. 
We implement all comparison models and \textsc{CompoSition} with an open source Fairseq toolkit~\cite{ott2019fairseq}. More details are provided in Appendix~\ref{sec:aes}.

\noindent\textbf{Evaluation Metrics.} For CoGnition, we use compound translation error rate (CTER~\cite{li2021compositional}) to measure the model’s ability of CG. Specifically, \textit{instance-level} CTER denotes the ratio of samples where the novel compounds are translated incorrectly, and \textit{aggregate-level} CTER denotes the ratio of samples where the novel compounds suffer at least one incorrect translation when aggregating all 5 contexts. To calculate CTER,~\citet{li2021compositional} manually construct a dictionary for all the atoms based on the training set, since each atom contains different translations. We also report character-level BLEU scores~\cite{DBLP:conf/acl/PapineniRWZ02} using SacreBLEU~\cite{DBLP:conf/wmt/Post18} as a supplement. For CFQ, we use exact match accuracy to evaluate model performance, where natural language utterances are mapped to meaning representations.

\subsection{Model Settings}
\label{sec:mst}

\begin{table}[!t]
    \centering
    \small
    \resizebox{1.0\linewidth}{!}{
    \begin{tabular}{@{}cc@{}c@{}c@{}c@{}}
      \toprule
     \multicolumn{1}{@{}c}{\bf Model}  & \multicolumn{1}{c}{\bf MCD1} & \multicolumn{1}{c}{\bf MCD2} & \multicolumn{1}{c}{\bf MCD3}& \multicolumn{1}{c@{}}{\bf Mean} \\
      \midrule
        LSTM+attention & 28.9 & 5.0 & 10.8 & 14.9 \\
        Transformer & 34.9 & 8.2 & 10.6 & 17.9 \\
        Universal Transformer & 37.4 & 8.1 & 11.3 & 18.9 \\
        Evolved Transformer & 42.4 & 9.3 & 10.8 & 20.8 \\
        CGPS & 13.2 & 1.6 & 6.6 & 7.1 \\
        NSEN & 5.1 & 0.9 & 2.3 & 2.8 \\
        T5-11B & 61.4 & 30.1 & 31.2 & 40.9 \\
        T5-11B-mod & 61.6 & 31.3 & 33.3 & 42.1 \\
        RoBERTa & 60.6 & 33.6 & 36.0 & 43.4 \\
        HPD & 72.0 & \bf 66.1 & \bf 63.9 & \bf 67.3 \\
        Dangle & \bf 78.3 & 59.5 & 60.4 & 66.1 \\
        RoBERTa+CReg & 74.8 & 53.3 & 58.3 & 62.1 \\
        \midrule
        \textsc{CompoSition} & 72.8 & 53.2 & 52.2 & 59.4 \\
       \bottomrule
    \end{tabular}
    }
    \caption{Exact-match accuracy on different MCD splits of CFQ. Results are averaged over 3 random runs.}
    \label{table:t2}
\end{table}
\noindent\textbf{Machine Translation.} We compare our method with previous competitive systems: 
(1) Transformer~\cite{vaswani2017attention}: first proposes a new encoder-decoder architecture based solely on attention mechanisms; (2) Transformer-Rela: only replaces sinusoidal (absolute) positional embedding with a relative one; (3) Transformer-Small: only decreases the number of encoder layers and decoder layers to 4, 4 respectively; (4) Transformer-Deep: only increases the number of encoder layers to 8; (5) Bow~\cite{raunak2019compositionality}: uses bag-of-words pre-training to improve the representation of the encoder upmost layer; (6) SeqMix~\cite{guo2020sequence}: synthesizes examples to encourage compositional behavior; (7) Dangle~\cite{zheng2021disentangled}: adaptively re-encodes (at each time step) the source input to disentangle the representation of the encoder upmost layer;\footnote{We use the same variant reported by~\citet{zheng2021disentangled} (i.e., Dangle-EncDec (abs)) with sinusoidal (absolute) positional embedding.} (8) Proto-Transformer~\cite{yin2022categorizing}: integrates prototypes of token representations over the training set into the source encoding to achieve the goal of categorization; (9) Transformer+CReg~\cite{yin2023consistency}: promotes representation consistency across samples and prediction consistency for a single sample; (10) $~$R-Dangle$_{\mathrm{sep}}$~\cite{zheng2022real}: disentangles their
representations and only re-encode keys periodically, at some interval; (11) DLCL~\cite{wang2019learning}: proposes an approach based on dynamic linear combination of layers (DLCL), and is one of the very popular EnocderFusion work. Our method is built on top of (1)-(4), i.e., \textsc{CompoSition}, \textsc{CompoSition}-Rela, \textsc{CompoSition}-Small and \textsc{CompoSition}-Deep. We also provide reasons for experiments on CoGnition without language models (see Appendix~\ref{sec:reclm}).

\noindent\textbf{Semantic Parsing.} We compare our method with previous competitive systems: (1) LSTM + attention: introduces attention mechanism~\cite{bahdanau2014neural} in LSTM~\cite{DBLP:journals/neco/HochreiterS97}; (2) Transformer~\cite{vaswani2017attention}; (3) Universal Transformer~\cite{DBLP:conf/iclr/DehghaniGVUK19}: combines the parallelizability and global receptive field of feed-forward
sequence models like the Transformer with the recurrent inductive bias of RNNs; (4) Evolved Transformer~\cite{DBLP:conf/icml/SoLL19}: uses wide depth-wise separable convolutions in the early layers of both the encoder and decoder; (5) CGPS~\cite{li2019compositional}: leverages prior knowledge of compositionality with two representations, and adds entropy regularization to the encodings; (6) NSEN~\cite{DBLP:conf/nips/FreivaldsOS19}: is derived from the Shuffle-Exchange network; (7) T5-11B~\cite{DBLP:journals/jmlr/RaffelSRLNMZLL20}: treats every natural language processing task as a text-to-text problem, and is therefore suitable for the semantic parsing tasks. T5-11B is a T5 model with 11B parameters finetuned on CFQ; (8) T5-11B-mod~\cite{furrer2020compositional}: shows that using masked language model (MLM) pre-training together with an intermediate representation leads to significant improvements in performance; (9) RoBERTa~\cite{liu2019roberta}: makes use of the RoBERTa-base model as the encoder and the randomly initialized Transformer decoder trained from scratch, where we use the same experimental settings of~\cite{zheng2021disentangled}; (10) HPD~\cite{NEURIPS2020_4d7e0d72}: proposes a novel hierarchical partially ordered set (poset) decoding paradigm, which consists of three components: sketch prediction, primitive prediction, and traversal path prediction; (11) Dangle~\cite{zheng2021disentangled}; (12) RoBERTa+CReg~\cite{yin2023consistency}; (13) \textsc{CompoSition}: builds on (9) with our method.

\begin{table}[!t]
\centering
\small
\resizebox{1.0\linewidth}{!}{
\begin{tabular}{lccll}
\toprule
\multirow{2}{*}{\bf Model} & \multicolumn{2}{c}{\bf Alleviate} & \multirow{2}{*} {\bf $~$CTER$_{\mathrm{Inst}}\downarrow$} & \multirow{2}{*}{\bf $~$CTER$_{\mathrm{Aggr}}\downarrow$}\\
& \bf E & \bf K, V \\
\midrule
Transformer & \xmark & \xmark & 28.4\% & 62.9\% \\
\midrule
\textsc{CompoSition} & \cmark & \xmark & 22.6\% (-5.8\%) & 55.1\% (-7.8\%) \\
\textsc{CompoSition} & \cmark & \cmark & \bf 20.4\% (-8.0\%) & \bf 52.0\% (-10.9\%) \\
\bottomrule
\end{tabular}
}
\caption{CTERs (\%) against alleviating E or K,V on the CG-test set, where $~$CTER$_{\mathrm{Inst}}$ and $~$CTER$_{\mathrm{Aggr}}$ denote instance-level and aggregate-level CTER respectively. E and K, V denote encoder and keys, values entanglement problem respectively.}
\label{table:t3}
\end{table}
\subsection{Results on CoGnition}
\label{sec:roc}

The main results on CoGnition are shown in Table~\ref{table:t1}. We observe that: \textbf{(1)} \textsc{CompoSition} gives \textbf{20.4\%}$~$CTER$_{\mathrm{Inst}}$ and \textbf{52.0\%}$~$CTER$_{\mathrm{Aggr}}$, with a significant improvement of \textbf{8.0\%} and \textbf{10.9\%} accordingly compared to the Transformer. Moreover, \textsc{CompoSition} significantly outperforms most baseline models under the almost same parameter settings,\footnote{We implement our approach based on Transformer-Deep for a fair comparison with Proto-Transformer.} indicating composing the syntactic and semantic information of sequences dynamically for a particular task is more beneficial to CG. Although Transformer+CReg achieves slightly better performance and contains fewer parameters, it is more complex and costly compared with \textsc{CompoSition}; \textbf{(2)} \textsc{CompoSition}, \textsc{CompoSition}-Rela, \textsc{CompoSition}-Small and \textsc{CompoSition}-Deep can deliver various performance improvements, demonstrating the general effectiveness of our method; \textbf{(3)} \textsc{CompoSition}-Deep performs better than Bow, Dangle and Proto-Transformer, indicating that focusing on alleviating the encoder entanglement problem only can achieve part of goals of CG as mentioned in Section~\ref{sec:intro}. 
% enhancing semantic information at token-level only can achieve part of goals of CG as mentioned in Section~\ref{sec:intro}. 
% For example, ``watch a movie'' should be translated as [\begin{CJK*}{UTF8}{gbsn}看电影\end{CJK*}] in most cases, which is not affected by the context to a certain extent. Bow, Dangle and Proto-Transformer can perform better than baseline models in this situation, however, for ``lost the dog he liked'' in Figure~\ref{fig:fig1}, the Chinese translation generated by they will most likely be [\begin{CJK*}{UTF8}{gbsn}丢失了狗他喜欢\end{CJK*}] instead of [\begin{CJK*}{UTF8}{gbsn}丢失了他喜欢的狗\end{CJK*}] due to ignoring syntactic information and its appropriate composition of both. 
% We also provide a case study in Section~\ref{sec:ecg} to further illustrate it. 
% It also suggests the reasonableness of keys, values entanglement problem we proposed to some extent, and \textsc{CompoSition} can mitigate this issue effectively. 
Compared to SeqMix, the improvement of \textsc{CompoSition} is more significant (2.3\% vs 10.9\%$~$CTER$_{\mathrm{Aggr}}$). SeqMix utilizes linear interpolation in the input embedding space to reduce representation sparsity, and we suppose that the  samples synthesized randomly may be unreasonable and harmful to model training; \textbf{(4)} It can be seen that Transformer is even slightly better than DLCL, indicating DLCL and \textsc{CompoSition} do not share the same motivation or scenario. 
% DLCL focuses on how to transform information across layers in deep neural network scenarios, while \textsc{CompoSition} is to learn to composing the syntactic and semantic representations of sequences dynamically for CG.
% ; \textbf{(4)} Bloom (7B) even obtains worse results (35.6\% vs 28.4\%$~$CTER$_{\mathrm{Inst}}$) than Transformer (35M). It indicates that large-scale language models, despite their huge number of parameters and training datasets, still fail to improve the CG performance.

\begin{figure}[!t]
    \centering
    \includegraphics[width=1.0\linewidth]{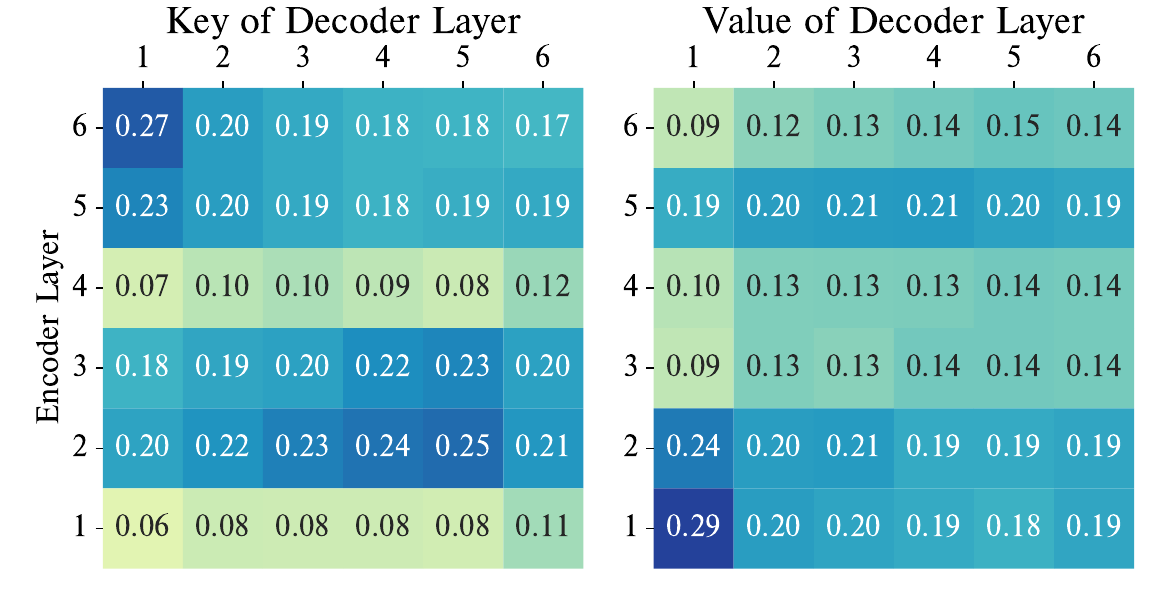}
    \caption{Learned composition weights (after normalized) that each encoder layer ($y$-axis) attending to keys or values of different decoder layers ($x$-axis).}
    \label{fig:fig4}
\end{figure}
\subsection{Results on CFQ}
\label{sec:rocfq}

The main results on CFQ are presented in Table~\ref{table:t2}. We observe that: \textbf{(1)} RoBERTa is comparable to T5-11B,  T5-11B-mod and outperforms other baseline systems without pre-training except HPD, indicating that pre-training indeed benefits CFQ;
% , which is consistent with the conclusions in~\citet{furrer2020compositional}; 
\textbf{(2)} \textsc{CompoSition} substantially boosts the performance of RoBERTa (\textbf{43.4 $\rightarrow$ 59.4}), about \textbf{37\%} relative improvements, and is in fact superior to T5-11B and T5-11B-mod. It also outperforms other baseline systems without pre-training except HPD. This result demonstrates that pre-training as a solution to CG also has limitations, and also indicates that \textsc{CompoSition} is complementary to pre-trained models; \textbf{(3)} HPD performs better than Dangle, RoBERTa+CReg and \textsc{CompoSition}, achieving 67.3 exact match accuracy, which is highly optimized for the CFQ dataset. On the contrary, \textsc{CompoSition}, RoBERTa+CReg and Dangle are generally applicable to any seq2seq models for solving any seq2seq tasks including MT, as mentioned in Section~\ref{sec:roc}. However, compared with competitive performance on CoGnition, the improvements brought by \textsc{CompoSition} is relatively moderate, and even worse than Dangle. The underlying reason is related to a recent finding that compositionality in natural language is much more complex than the rigid, arithmetic-like operations~\cite{li2021compositional,zheng2021disentangled,DBLP:conf/acl/DankersBH22}. MT is paradigmatically close to the tasks typically considered for testing compositionality in natural language, while our approach is more suitable for dealing with such scenarios.

\begin{table}[!t]
    \centering
    \small
    \resizebox{1.0\linewidth}{!}{
    \begin{tabular}{cll}
        \toprule
        \bf Model & \bf $~$CTER$_{\mathrm{Inst}}\downarrow$ & \bf $~$CTER$_{\mathrm{Aggr}}\downarrow$ \\
        \midrule
        Transformer & 28.4\% & 62.9\% \\
        \textsc{CompoSition}-SA & 22.2\% (-6.2\%) & 53.8\% (-9.1\%) \\
        \textsc{CompoSition}-FF & 22.6\% (-5.8\%) & 55.6\% (-7.3\%) \\
        \textsc{CompoSition}-SA \& FF & \bf 20.4\% (-8.0\%) & \bf 52.0\% (-10.9\%) \\
        \bottomrule
    \end{tabular}
    }
    \caption{CTERs (\%) against composing different source information on the CG-test set.}
    \label{table:t4}
\end{table}
\section{Analysis}
\label{sec:ana}

In this section, we conduct in-depth analyses of \textsc{CompoSition} to provide a comprehensive understanding of the individual contributions of each component. For all experiments, we train a \textsc{CompoSition} (6-6 encoder and decoder layers) instead of other experimental settings on the CoGnition dataset, unless otherwise specified.

\subsection{Effects of Specific Keys and Values of Different Decoder Layers}
\label{sec:kvep}

As mentioned in Section~\ref{sec:intro} and~\ref{sec:cmp}, we hypothesize that keys, values entanglement problem exists.\footnote{It is noteworthy that the representation of the encoder upmost layer serves as the same key and value passing into every decoder layer in the Transformer.}
% So, we only visualize it once in Fig.~\ref{fig:fig5}.\label{fn:fn5}
% source keys and values representations passing into different decoder layers are also entangled (keys, values entanglement problem), while the representation of the encoder upmost layer serves as the same key and value passing into every decoder layer in the Transformer. 
It is clear that our hypothesized keys, values entanglement problem is an extension to encoder entanglement problem. We show curiosity about whether this problem exists, and \textsc{CompoSition} can alleviate it. In this experiment, we investigate its influence on CoGnition. As shown in Table~\ref{table:t3}, we observe certain improvements (\textbf{-5.8\%} and \textbf{-8.0\%}$~$CTER$_{\mathrm{Inst}}$, \textbf{-7.8\%} and \textbf{-10.9\%}$~$CTER$_{\mathrm{Aggr}}$) when separately alleviating the encoder or keys, values entanglement problem.\footnote{We use one or 2$N$ learnable vectors to generate one or 2$N$ representations passing into $N$ decoder layers.} It suggests that our method can alleviate both problems separately, and learning to compose information of different encoder layers dynamically can improve CG performance. Furthermore, the improvement brought from alleviating keys, values entanglement problem is more significant than that brought from alleviating encoder entanglement problem (52.0\% vs 55.1\%$~$CTER$_{\mathrm{Aggr}}$), demonstrating the reasonableness of keys, values entanglement problem.

% \begin{table}[!t]
%     \centering
%     \small
%     \resizebox{1.0\linewidth}{!}{
%     \begin{tabular}{c|c|c|c}
%         \hline
%         \bf Model  & \bf GPU Memory & \bf upd/s & \bf token/s \\
%         \hline
%         \hline
%         Transformer & 9.4G(1.00$\times$) & 4.71(1.00$\times$) & 5986.00(1.00$\times$) \\
%         \hline
%         \hline
%         Bow & 9.4G(1.00$\times$) & - & 5936.10(0.99$\times$) \\
%         Dangle & 63.3G(6.73$\times$) & 1.23(0.26$\times$) & 2505.58(0.42$\times$) \\
%         Proto-Transformer & 10.2G(1.09$\times$) & - & 5638.20(0.94$\times$) \\
%         \hline
%         \hline
%         \textsc{CompoSition} & 8.8G(0.94$\times$) & 4.05(0.86$\times$) & 5728.01(0.97$\times$)  \\
%         \hline
%     \end{tabular}
%     }
%     \caption{Computational and memory cost on the CG-test set. ``-'' means ``not applicable'' since both Bow and Proto-Transformer contain two-stage training process. ``upd/s'' and ``token/s'' are abbreviated notations for ``training updates per second'' and ``generated tokens per second''.}
%     \label{table:t5}
% \end{table}
% \begin{figure}[!t]
%     \centering
%     \includegraphics[width=0.80\linewidth]{figures/vis.pdf}
%     \caption{PCA visualization of hidden states for three randomly selected examples of CG-test set. DL denotes Decoder Layer. Circle markers denote the hidden states of the Transformer encoder, while other different shapes of markers correspond to source keys and values of different decoder layers in \textsc{CompoSition}.}
%     \label{fig:fig5}
% \end{figure}
\begin{figure}[!t]
    \centering
    \includegraphics[width=1.0\linewidth]{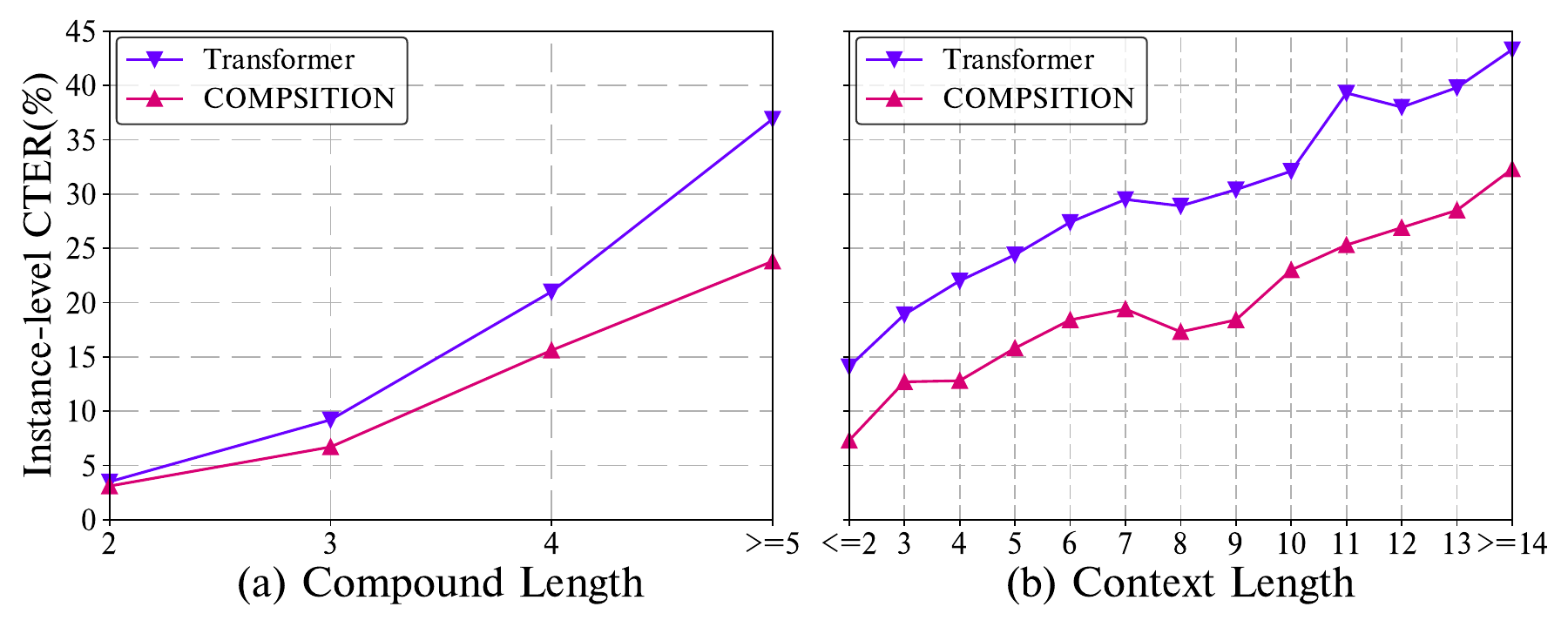}
    \caption{$~$CTER$_{\mathrm{Inst}}$ of \textsc{CompoSition} and Transformer over the different compound and context lengths.}
    \label{fig:fig6}
\end{figure}
To further illustrate the reasonableness of keys, values entanglement problem and understand how \textsc{CompoSition} alleviates it, we visualize the learned composition weights of \textsc{CompoSition} after normalized.\footnote{We only use representations outputted by Eq.~\ref{eq6} for brevity.} Specifically, we train \textsc{CompoSition} on CoGnition and then extract $W_{k}^{i}, W_{v}^{i}$ (see Section~\ref{sec:plc}) to visualize them. Ideally, 
% \textsc{CompoSition} should generate different keys and values representations for each decoder layer. In other words, 
each key or value of different decoder layers should pay different attention weights to different encoder layers’
information. As shown in Figure~\ref{fig:fig4}, the learned composition weights (after normalized) are mutually distinct for keys and values of different decoder layers, which implies \textsc{CompoSition} learns different dynamic composition modes for keys and values of every decoder layer respectively. In addition, it also indicates the reasonableness of keys, values entanglement problem we proposed, since keys and values of different decoder layers utilize more than just the information of the encoder topmost layer. More importantly, it also emphasizes our method provides an effective composition of syntactic and semantic information, i.e., a specific combination for a particular task. 
To further demonstrate it, we also provide a toy experiment in Appendix~\ref{sec:appacssi}.
% that \textsc{CompoSition} can learn to compose information from all encoder sub-layers \textbf{appropriately} to generate different keys and values passing into different decoder
% layers. To further demonstrate the importance of \textbf{appropriateness}, we also provide a toy experiment on CoGnition in Appendix~\ref{sec:appacssi}.

% We also visualize the hidden states for Transformer and \textsc{CompoSition} to support the above conclusions. Specifically, we train both models on CoGnition and test on three randomly selected examples of CG-test set. Then, we extract the hidden states of source keys and values passing into the 1st, 4th, and 6th decoder layer\footref{fn:fn5}
% % \footnote{The same source representation is used to predict all target symbols in Transformer. So, we only visualize it once.}
% and use Principal Component Analysis~\cite{bro2014principal} (PCA) to visualize them. As shown in Fig.~\ref{fig:fig5}, in contrast to the Transformer, \textsc{CompoSition} obtains specialized keys and values representations passing into different decoder layers, suggesting that keys, values entanglement problem indeed exists.

\begin{figure}[!t]
    \centering
    \includegraphics[width=0.9\linewidth]{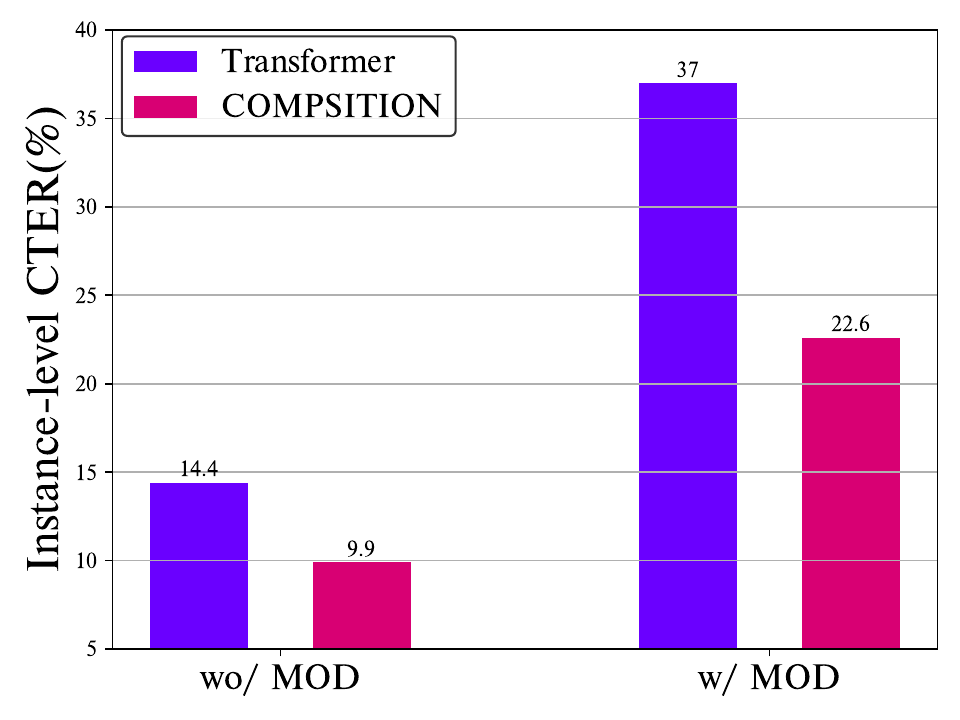}
    \caption{$~$CTER$_{\mathrm{Inst}}$ on compounds w/o and w/ MOD.}
    \label{fig:fig7}
\end{figure}
\subsection{Effects of Composing Information of Encoder Layers or Sub-layers}
\label{sec:elsi}

As mentioned in Section~\ref{sec:met}, the Transformer encoder layer consists of two sub-layers. We assume that sub-layers may contain language information in different aspects, which may produce better generalization results. Therefore, we are curious about whether composing different encoder layers’ or sub-layers’ information is more beneficial to CG.
% , and which sub-layer's information affects the model’s ability of CG to a greater extent. 
In this experiment, we investigate its influence on CoGnition. Specifically, we train \textsc{CompoSition} to compose representations outputted by either Eq.~\ref{eq5} or~\ref{eq6} or a combination of both dynamically. Results are presented in Table~\ref{table:t4}. We observe certain improvements (\textbf{-6.2\%} and \textbf{-5.8\%}$~$CTER$_{\mathrm{Inst}}$) when separately composing SA- and FF-level representations, where SA and FF denote representations outputted by Eq.~\ref{eq5} and~\ref{eq6} respectively. Furthermore, the combination of both them brings further improvement (\textbf{-8.0\%}$~$CTER$_{\mathrm{Inst}}$), which illustrates that the information in different encoder sub-layers is complementary and has cumulative gains. It also suggests that syntactic and semantic information brought by SA or FF is similar, but slightly different~\cite{li2020shallow}, and can improve generalization performance respectively. It can be seen that the results of \textsc{CompoSition}-SA and \textsc{CompoSition}-FF presented in Table~\ref{table:t4} are basically the same, and the improvements brought by the combination of both them is relatively moderate. 
% In addition, it is clear that \textsc{CompoSition}-FF is slightly better than \textsc{CompoSition}-SA. The underlying reason we consider is that the representation after passing into the feed-forward sub-layer is more abstract, which instead lacks part of the relevant information, leading even a little performance degradation.

% \subsection{Computational and Memory Cost}
% \label{sec:cmc}

% As mentioned in Section~\ref{sec:cmp}, Our method is simple and applicable to a wide range of more realistic tasks due to little increase in computational and memory cost compared to the Transformer baseline, which is in contrast to~\cite{raunak2019compositionality,zheng2021disentangled,yin2022categorizing}. In this experiment, we investigate its influence on CoGnition. Results are listed in Table~\ref{table:t5}. We observe \textsc{CompoSition} keeps almost the same training and decoding speed as Transformer, and requires even less GPU memory than Transformer during training. On the contrary, both Bow and Proto-Transformer bring huge training costs due to their complicated two-stage training process. In addition, although Dangle is a simple and effective one-stage method, it prohibitively increase the training and decoding cost. Thus, it is clear that our method achieves promising results while brings little increase in computational and memory cost compared with Transformer. 

\begin{table*}
    \centering
    \small
    \begin{CJK*}{UTF8}{gbsn}
    \resizebox{1.0\linewidth}{!}{
    \begin{tabular}{p{3.6cm}<{\centering}|p{6.0cm}<{\centering}|p{7.0cm}<{\centering}}
    \toprule
    \textbf{Source} & \textbf{Transformer} & \textbf{\textsc{CompoSition}} \\
    \hline
    \textbf{The waiter he liked} & 他喜欢穿对方的衣服 。 & 他喜欢的服务员穿着彼此的衣服。\\
    wore each other's clothes.  & (\textbf{He liked} to wear each other's clothes.) & (\textbf{The waiter he liked} wore each other's clothes.) \\
    \midrule
    \textbf{The waiter he liked} came & 服务员来了，把那个恶霸赶走了。 & 他喜欢的服务员过来把那个恶霸赶走了。\\
    by and chased the bully off.  & (\textbf{The waiter} came by and chased the bully off.)  & (\textbf{The waiter he liked} came by and chased the bully off.)\\
    \midrule
    \textbf{The waiter he liked} & 服务员喜欢拿起他的邮件。& 他喜欢的服务员拿起了他的邮件。  \\ 
    picked up his mail. & (\textbf{The waiter liked} to pick up his mail.) & (\textbf{The waiter he liked} picked up his mail.) \\
    \bottomrule
    \end{tabular}
    }
    \caption{
    Example translations of Transformer vs \textsc{CompoSition}.
    \textbf{The bold characters} denote the novel compounds and corresponding translations.
    }
    \label{table:t5}
    \end{CJK*}
\end{table*}
\subsection{Effects on Compositional Generalization}
\label{sec:ecg}

\noindent\textbf{Compound Length and Context Length.} Longer compounds have more complex semantic information and longer contexts are harder to comprehend, making them more difficult to generalize~\cite{li2021compositional}. We classify the test samples by compound length and context length, and calculate the $~$CTER$_{\mathrm{Inst}}$. In Figure~\ref{fig:fig6}, we can observe that \textsc{CompoSition} generalizes better as the compound and context grows longer compared to Transformer. In particular, \textsc{CompoSition} gives a lower CTER by \textbf{11.0\%} over samples when the context length is more longer than 13 tokens. It suggests that our approach can better captures the compositional structure of human language.

\noindent\textbf{Complex Modifier.} The postpositive modifier atom (MOD) is used to enrich the information of its preceding word (e.g., \emph{he liked} in the phrase \emph{lost the dog he liked}), which is challenging to translate due to word reordering from English to Chinese. We divide the test samples into two groups according to compounds with (w/) or without (wo/) MOD. In Figure~\ref{fig:fig7}, we observe that the advantage of \textsc{CompoSition} grows larger in translating the compounds with MOD, demonstrating its superiority in processing complex semantic composition. 

\noindent\textbf{Case Study.} We present 3 source examples containing a novel compound \textit{the waiter he liked} with MOD and 4 atoms, and their translations in Table~\ref{table:t5}. For all samples, correct translations denote that the novel compounds are translated correctly. \textsc{CompoSition} correctly translates the novel compounds across different contexts for all samples, while Transformer suffers from omitting different atoms. 
% In the first example, the translation of \emph{the waiter} is omitted by Transformer. Similar mistakes can be observed in the second and third examples, where Transformer omits \emph{he liked} and \emph{he} respectively.
For example, the translation of \emph{the waiter} is omitted in the first example, \emph{he liked} is omitted in the second example and \emph{he} is omitted in the third example. 
Our results not only contain the correct compound translations but also achieve better translation quality, while Transformer makes errors on unseen compositions, confirming the necessity of composing the syntactic and semantic representations of sequences dynamically.

% \subsection{Effects of the Number of Encoder Layers}
% \label{sec:nel}

% As mentioned in Section~\ref{sec:intro}, different layers of the Transformer encoder contain language information in different aspects. We assume that 

\section{Conclusion}
\label{sec:conl}

% In this paper, we examine CG from a new perspective. Starting from recent studies on analyzing how Transformer perform, we find that the representation of the Transformer encoder upmost layer contains less syntactic information. Meanwhile, we hypothesize that source keys and values representations passing into different decoder layers are also entangled. Therefore, inspired by insights from human strategies for CG, 
% In this paper, we propose \textsc{CompoSition}, an extension to Seq2Seq models to compose different encoder sub-layers' representations appropriately for generating specialized keys and values passing into different decoder layers via introducing a composed layer between the encoder and decoder.
% learn to compose the syntactic and semantic representations of sequences appropriately for solving keys, values entanglement problem. 
% Specifically, we introduce a composed layer between the encoder and decoder to compose representations of different encoder layers appropriately for generating different keys and values passing into different decoder layers. 
In this paper, we examine CG from a new perspective, i.e., utilizing
different encoder layers’ information. Specifically, we propose an extension to seq2seq models which composes different encoder layers' representations dynamically to generate specific keys and values passing into different decoder layers. 
Experiments on CoGnition and CFQ have shown the effectiveness of our proposal on CG without any dataset or task-specific modification. To our knowledge, we are the first to point out a new representation entanglement problem and investigate how to utilize information of different encoder layers benefits CG, achieving promising results on two realistic benchmarks. We hope the work and perspective presented in this paper can inspire future related work on CG. 

\section*{Limitations}
There are two limitations of our approach. Firstly, compared
with competitive performance on CoGnition, the improvements brought by \textsc{CompoSition} on CFQ is relatively moderate, and even worse than some competitive methods. Hence, \textsc{CompoSition} is more suitable for tasks typically considered for testing compositionality in natural language. We strongly recommend researchers pay more attention to tasks evaluating compositionality on natural language. Meanwhile, we regard that designing a more general method that can improve generalization performance in both synthetic and natural scenarios is a promising direction to explore in the future. Secondly, our method is mostly applicable to any seq2seq models which adopt an encoder-decoder architecture instead of encoder-only or decoder-only architecture. However, the methodology of the proposed \textsc{CompoSition} is still rather general to any seq2seq models which adopt any architecture, since we can use the randomly initialized encoder or decoder to constitute the encoder-decoder architecture.
% However, our experiment result on CFQ shows that \textsc{CompoSition} slightly hurts generalization performance (in comparison with other state-of-the-art methods). We strongly suggest everyone should pay more attention to tasks using real data to evaluate compositionality on natural language. Hence, we regard that designing a more general 

\section*{Acknowledgement}
We thank all the anonymous reviewers for their insightful and valuable comments. This work is supported by National key R\&D Program of China (Grant no.2022ZD0116101), the Key Support Project of NSFC-Liaoning Joint Foundation (Grant no. U1908216), and the Project of Research and Development for Neural Machine Translation Models between Cantonese and Mandarin (No. WT135-76).

\bibliography{custom}
\bibliographystyle{acl_natbib}

\appendix

\section{Preliminary Analysis}
\label{sec:apa}

In this section, we analyze the amount of syntactic and semantic information captured by different encoder layers in the Transformer under MT scenarios. We aim at analyzing the representations learned by different encoder layers of different models through probing the encoder as input representation for various prediction tasks. We measure the importance of input features for various tasks by evaluating the ability of the decoder. Specifically, we use a fixed encoder representation as input and two different tasks, i.e., Part-of-Speech (POS) tagging, and Semantic tagging, to evaluate the syntactic and semantic information contained in different encoder layers respectively. The reason is that we assume if the input representation effectively captures a property (syntactic or semantic information), then the decoder can easily predict that property.

\begin{figure}[!t]
    \centering
    \includegraphics[width=1.0\linewidth]{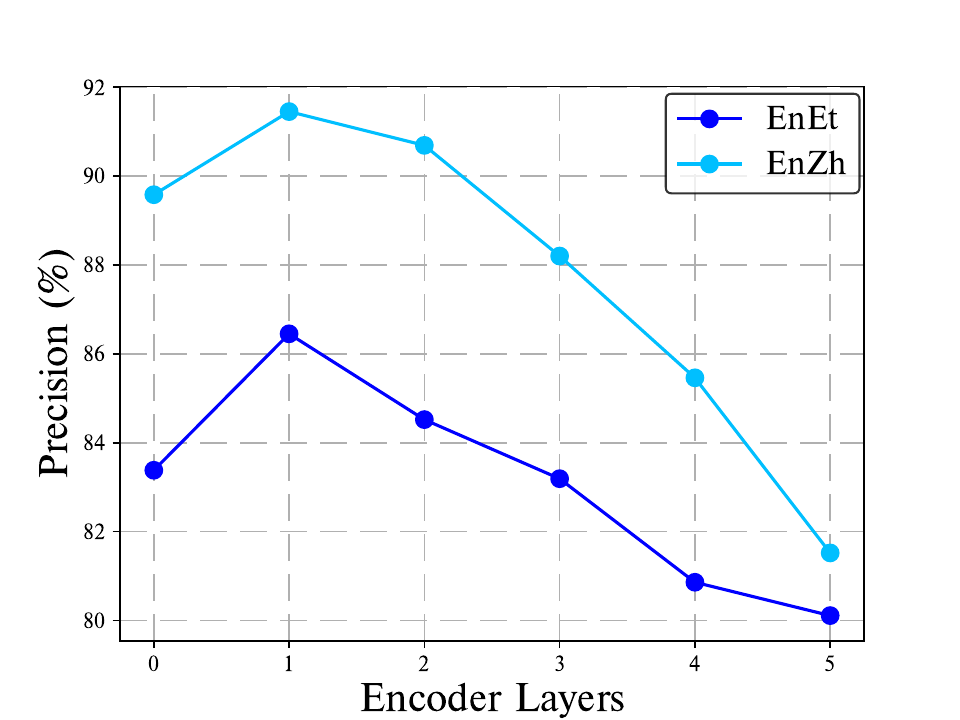}
    \caption{Precision (\%) against different encoder layers' representations as input on the test set of POS tagging task.}
    \label{fig:fig8}
\end{figure}
To explore the precise effects of information captured by different encoder layers, we train the Transformer on the WMT18 English $\rightarrow$ Chinese (EnZh, rich-resource), English $\rightarrow$ Estonian (EnEt, low-resource)\footnote{We use two datasets with different sizes to analyze information captured by different encoder layers across models with different translation quality and target language.} by following the same settings of~\citet{raganato2018analysis}.\footnote{The provided datasets are freely available at~\url{https://www. statmt.org/wmt18/translation-task.html}.} After training the MT models, we freeze the encoder parameters, and only train one decoder layer\footnote{We follow the same settings of~\citet{raganato2018analysis} and adopt one attention head and one feed-forward sub-layer to consitute the decoder layer.} for each task, since we expect the decoder should not have overly significant impact on the model's performance of different tasks. We then analyze the amount of syntactic and semantic information in different encoder layers via evaluating the different encoder layers' performance of corresponding task. We use the Universal Dependencies English Web Treebank v2.0~\cite{ZemanPSHNGLPPPT17} for POS tagging (syntactic task) and the annotated data from the Parallel Meaning Bank (PMB)~\cite{BosEBAHNLN17} for Semantic tagging (semantic task).\footnote{We follow the same data preprocess process of~\citet{ZemanPSHNGLPPPT17} and~\citet{BosEBAHNLN17}.} We use precision to evaluate model performance.

\begin{figure}[!t]
    \centering
    \includegraphics[width=1.0\linewidth]{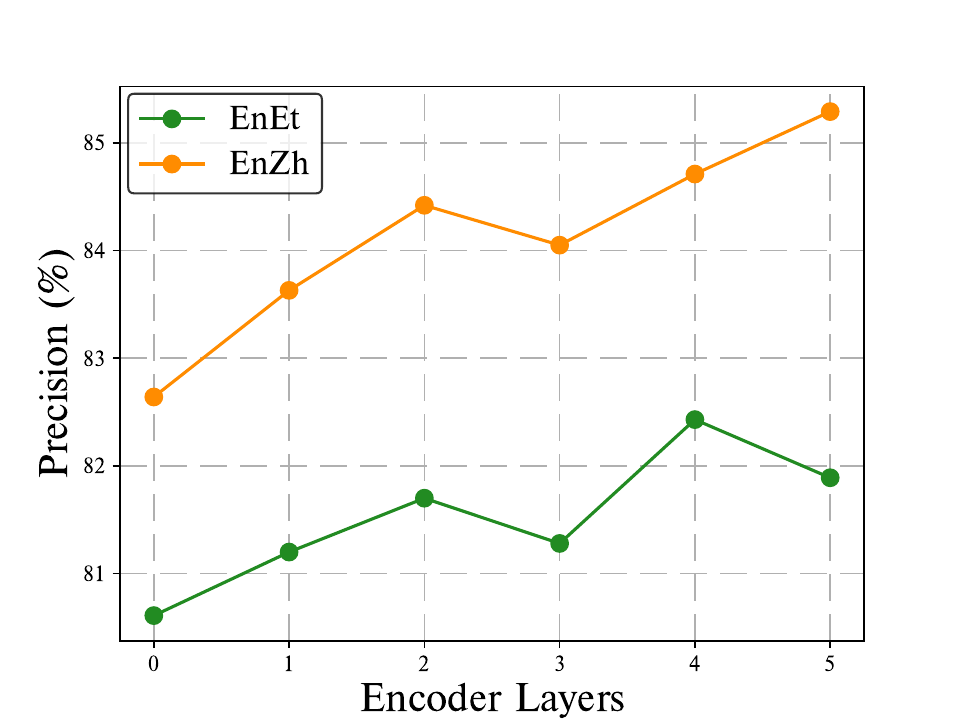}
    \caption{Precision (\%) against different encoder layers' representations as input on the test set of Semantic tagging task.}
    \label{fig:fig9}
\end{figure}
Results on POS tagging and Semantic tagging are presented in Figure~\ref{fig:fig8} and~\ref{fig:fig9} respectively. We observe that:
\begin{itemize}
    \item For EnEt and EnZh,
    % the representations of the first 3 layers as input perform better than that of the last 3 layers as input on the POS tagging task, and 
    the performance tends to decrease as the number of layers increase.
    \item For EnEt and EnZh,
    % the representations of the last 3 layers as input perform better than that of the first 3 layers as input on the Semantic tagging task, and 
    the performance tends to increase as the number of layers increase.
\end{itemize}
% It is clear that the syntax information is encoded mostly in the first 3 layers, while the semantic information is encoded mostly in the last 3 layers.
Therefore, we can conclude that \textbf{the bottom layers of the Transformer encoder contain more syntactic information and the top ones contain more semantic information}, and the information encoded by each encoder layer transforms from syntactic to semantic as the number of layers increase.
% \begin{table}[!t]
%     \centering
%     \small
%     \begin{tabular}{c|l|l}
%         \hline
%         \textbf{Encoder Layers} & \textbf{En $\rightarrow$ Et} & \textbf{En $\rightarrow$ Zh} \\
%         \hline
%         0 & 84.49 & 90.81 \\
%         0 & 84.49 & 90.81 \\
%         \hline
%     \end{tabular}
%     \caption{xx.}
%     \label{table:t8}
% \end{table}
\section{Experimental Settings}
\label{sec:aes}

For CoGnition, we set hidden size to 512 and feed-forward dimension to 1,024. The number of encoder and decoder layers are 6, 6 and the number of attention heads are 4. The model parameters are optimized by Adam \cite{adam}, with $\beta_{1} = 0.9$, $\beta_{2} = 0.98$. The learning rate is set to 5e-4 and the number of warm-steps is 4000. We set max tokens as 8,192 tokens for iteration. We use one GeForce GTX 2080Ti for training with 100,000 steps and decoding. We report the average performance over 6 random seeds provided in~\citet{li2021compositional}. We train all \textsc{CompoSition} models from scratch.
For CFQ, we use the base RoBERTa with 12 encoder layers, which is combined with a Transformer decoder that has 2 decoder layers with hidden size 256 and feed-forward dimension 512. We use a separate target vocabulary. The number of attention heads are 8. The model parameters are optimized by Adam \cite{adam}, with $\beta_{1} = 0.9$, $\beta_{2} = 0.98$. The learning rate is set to 1e-4 and the number of warm-steps is 4000. We set max tokens as 4,096 tokens for iteration. We use one GeForce GTX 2080Ti for training with 45,000 steps and decoding. We report the average performance over 3 random seeds provided in~\citet{zheng2021disentangled}. We train \textsc{CompoSition} built on top of RoBERTa with full parameter fine-tuning.

\begin{table}[!t]
    \centering
    \small
    \resizebox{1.0\linewidth}{!}{
    \begin{tabular}{cll}
        \toprule
        \textbf{Model} & \textbf{$~$CTER$_{\mathrm{Inst}}\downarrow$} & \textbf{$~$CTER$_{\mathrm{Aggr}}\downarrow$} \\
        \midrule
        Transformer & 28.4\% & 62.9\% \\
        Transformer-accu & failed & failed \\
        \textsc{CompoSition} & \textbf{20.4\% (-8.0\%)} & \textbf{52.0\% (-10.9\%)} \\
        \bottomrule
    \end{tabular}
    }
    \caption{CTERs (\%) against Transformer-accu vs \textsc{CompoSition} on the CG-test set.}
    \label{table:t6}
\end{table}

\section{Effects of the Effective Composition}
\label{sec:appacssi}

As mentioned in Section~\ref{sec:met}, we introduce the composed layer between the encoder and decoder to compose different encoder sub-layers' information dynamically to generate specific keys and values passing into different decoder layers. We show curiosity about whether the composed layer can fuse all encoder sub-layers' information effectively. Therefore, we conduct a toy experiment on CoGnition. Specifically, all encoder sub-layers' information is accumulated to serve as the same key and value passing into every decoder layer (called Transformer-accu),\footnote{$\{y_{0}, ..., y_{l} \}$ are the output of the encoder layers $0 \sim l$. The input of keys and values of decoder layer $i$ is $x_{i} = y_{0} + \cdots + y_{i-1}$, where $0 < i < L$.} rather than composing them dynamically like we do. Results are listed in Table~\ref{table:t6}. Transformer-accu even fails to train. It suggests that even if the syntactic and semantic information of sequences is considered, the inappropriate combinations will instead bring noise to significantly affect the model's CG performance.

\section{Effects of Representations from Low-layer Encoder}
\label{sec:eler}

To verify the low-layer encoder representations are also essential to our approach, we only evaluate our approach on CoGnition with the collected encoder representations of the top three layers. Results are presented in Table~\ref{table:t7}. We can observe that only composing the representations of the top three encoder layers leads to a sharp drop in performance (27.0\% vs 20.4\%$~$CTER$_{\mathrm{Inst}}$), but still outperforms the Transformer baseline (27.0\% vs 28.4\%$~$CTER$_{\mathrm{Inst}}$). It further demonstrates the distinct difference between our method and the findings introduced by previous studies on EncoderFusion. It also reflects our starting point is correct, i.e., exploring how to compose syntactic and semantic information. It can be seen that \textsc{CompoSition}'s performance is dramatically reduced given only semantic information (the last three encoder layers' information).
\begin{table}[!t]
    \centering
    \small
    \resizebox{1.0\linewidth}{!}{
    \begin{tabular}{lll}
        \toprule
        \bf Model & \bf $~$CTER$_{\mathrm{Inst}}\downarrow$ & \bf $~$CTER$_{\mathrm{Aggr}}\downarrow$ \\
        \midrule
        Transformer & 28.4\% & 62.9\% \\
        \textsc{CompoSition}-Half & 27.0\% (-1.4\%) & 61.3\% (-1.6\%) \\
        \textsc{CompoSition} & \bf 20.4\% (-8.0\%) & \bf 52.0\% (-10.9\%) \\
        \bottomrule
    \end{tabular}
    }
    \caption{CTERs (\%) against Transformer, \textsc{CompoSition} and \textsc{CompoSition}-Half on the CG-test set.}
    \label{table:t7}
\end{table}

\section{Reasons for Experiments on CoGnition without Language Models}
\label{sec:reclm}

We do not conduct experiments on CoGnition with language models for two reasons. \textbf{First}, CoGnition is constructed to test CG performance in MT scenarios with simple sentence pairs (see Figure~\ref{fig:fig3}), however, language models are trained on vast amounts of multilingual sentences or bilingual sentence pairs. It is contrary to the compositional generalization task itself, since we can not guarantee that every sentence in the test set is a novel combination from known components for language models. \textbf{Second}, it is unfair to compare large language models with systems without pre-training. We strongly recommend researchers pay more attention to conduct experiments on CoGniton without language models.

% \usepackage{multirow}

% \usepackage{multirow}

\end{document}